\documentclass[runningheads]{llncs}
\usepackage{graphicx}
\usepackage{multirow}
\usepackage{makecell}
\usepackage{color}
\usepackage{graphicx}
\usepackage{amsmath}
\usepackage{booktabs}
\usepackage{subfig}
\usepackage{float}
\usepackage{listings}
\begin{document}
%
%\title{Enriching biomedical datasets with negative statements}
\title{Benchmark datasets for biomedical knowledge graphs with negative statements}
\author{Rita T. Sousa \and
Sara Silva \and Catia Pesquita}
\authorrunning{Sousa et al.}
\institute{LASIGE, Faculdade de Ci\^{e}ncias da Universidade de Lisboa\\
\email{\{risousa,sgsilva,clpesquita\}@ciencias.ulisboa.pt}}
\maketitle              % typeset the header of the contribution
\begin{abstract}
Knowledge graphs represent facts about real-world entities. Most of these facts are defined as positive statements. 
The negative statements are scarce but highly relevant under the open-world assumption. Furthermore, they have been demonstrated to improve the performance of several applications, namely in the biomedical domain.
However, no benchmark dataset supports the evaluation of the methods that consider these negative statements. 

We present a collection of datasets for three relation prediction tasks - protein-protein interaction prediction, gene-disease association prediction and disease prediction - that aim at circumventing the difficulties in building benchmarks for knowledge graphs with negative statements. These datasets include data from two successful biomedical ontologies, Gene Ontology and Human Phenotype Ontology, enriched with negative statements. 

%We also provide knowledge graph embeddings for each task with two popular path-based methods. 
We also generate knowledge graph embeddings for each dataset with two popular path-based methods and evaluate the performance in each task. The results show that the negative statements can improve the performance of knowledge graph embeddings.

\end{abstract}

%%%%%%%%%%%%%%%%%%%%%%%%%%%%%%%%%%%%%%%%%%%%%%%%%%%%%%% INTRODUCTION
\section{Introduction}

\begin{figure}[!tb]
    \centering
    \includegraphics[width=0.5\textwidth]{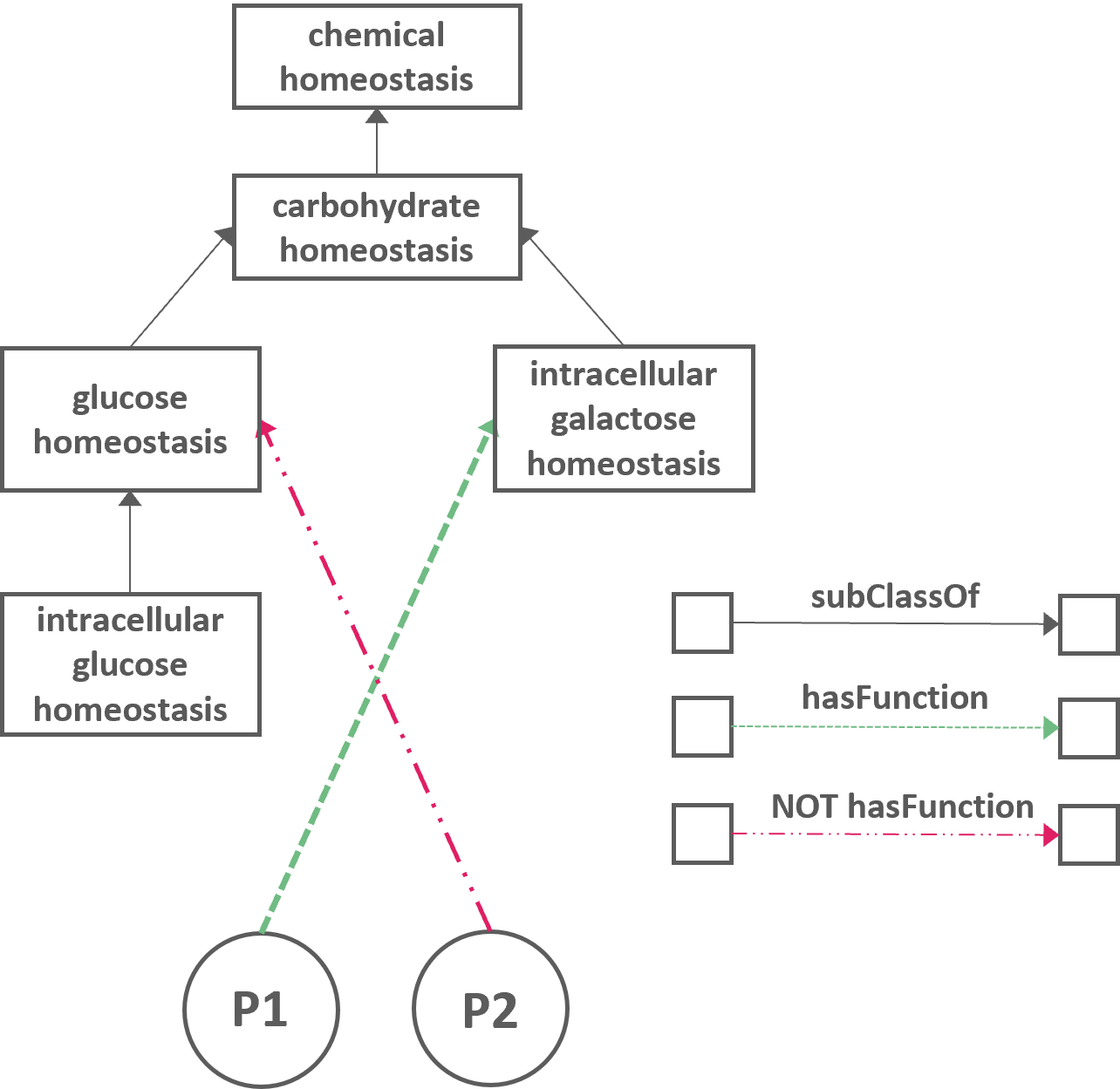}
    \caption{A GO KG subgraph with positive and negative statements describing two proteins.}
    \label{fig:kg}
\end{figure}

%Positive and Negative statements
Knowledge Graphs (KGs) have been used to represent knowledge about real-world entities and their relationships.
Most KGs use ontologies as a backbone to describe entities through ontology-based annotation, which associates an entity with a class.
% Given their semantic richness, they are widely used as background knowledge in various applications, namely in the biomedical domain.
These annotations are commonly represented as positive statements establishing that an ontology class describes an entity. For example, in the biomedical domain, positive statements express that a protein $P1$ performs $intracellular\_galactose\_homeostasis$ as defined in the Gene Ontology (GO)~\cite{GO2018}. Negative statements are extremely rare but can be used to declare that a given protein $P2$ does not perform $glucose\_homeostasis$ (Figure~\ref{fig:kg}).  

%Importance of negative statements
The lack of negative statements is a significant issue because KGs operate under the open-world assumption. Therefore, this lack of information can lead to confusion regarding whether the absence of a positive statement is due to a lack of knowledge or the actual absence of the relationship.
Moreover, the importance of negative statements to produce more accurate representations of entities in a KG~\cite{liu2021computational,gaudet2017gene} and improving performance in different applications~\cite{fu_neggoa_2016,vesztrocy2020benchmarking} is increasingly recognized in the biomedical domain.

% Why we need our datasets
While there have been attempts to enhance current KGs with interesting negative statements, to the best of our knowledge, no benchmark datasets have been established to evaluate learning tasks over those KGs. 
With this in mind, we enrich existing biomedical KGs with negative statements and propose a collection of datasets for different biomedical tasks of relation prediction.
The biomedical domain was selected because biomedical KGs are usually back-boned by biomedical ontologies that can express negation. Additionally, negative statements have been considered relevant for different biomedical applications~\cite{kulmanov2021semantic}. 
Our datasets are grouped according to the task: protein-protein interaction (PPI) prediction, gene-disease association (GDA) prediction and disease prediction. 
Regarding the KGs, we enrich two successful biomedical ontologies: GO which covers distinct semantic aspects of gene products' function, and Human Phenotype Ontology (HP) which describes the universe of concepts related to phenotypic abnormalities found in human hereditary diseases. 
% However, having benchmark datasets is fundamental for evaluating and comparing the performance of machine learning models that work over these KGs.

%%%%%%%%%%%%%%%%%%%%%%%%%%%%%%%%%%%%%%%%%%%%%%%%%%%%%%% RELATED WORK
\section{Related Work}

Several approaches to enriching existing KGs with interesting negative statements have been proposed.
Arnaout \textit{et al.}~\cite{arnaout_negative_2021-1} proposed a method to enrich Wikidata by including interesting negative statements, which led to improvements in tasks involving entity summarization and decision-making.

In the biomedical domain, several approaches tackle the lack of negative statements in biomedical ontologies, such as GO. 
The number of functions that a protein does not have is larger than the number of functions it has. Therefore, the number of negative statements describing proteins in the GO should be several orders of magnitude greater than the number of positive statements.
Youngs \textit{et al.}~\cite{youngs2014NoGO} designed two algorithms to predict negative statements for GO and populate the NoGo database, one based on empirical conditional probability and the other on topic modeling applied to genes and annotation. 
Fu \textit{et al.}~\cite{fu_neggoa_2016} introduced NegGOA, a new method to enrich the GO with relevant negative statements indicating that a protein does not perform a given function. This method exploits the GO by using hierarchical semantic similarity between GO terms. The enriched GO was used for protein function prediction. 
Later, Vesztrocy \textit{et al.}~\cite{vesztrocy2020benchmarking} presented a benchmark based on a balanced test set of positive and negative statements. The negative statements are generated from expert-curated annotations of protein families on phylogenetic trees. The results of this work demonstrated that negative statements improve protein function prediction.  
Regarding the HP, although the importance of negative statements in gene-phenotype prediction is recognized, the enrichment with negative statements has yet to be investigated~\cite{liu2021computational}.

%%%%%%%%%%%%%%%%%%%%%%%%%%%%%%%%%%%%%%%%%%%%%%%%%%%%%%% BUILDING THE DATASETS
\section{Building the Datasets}
We present a collection of datasets that work over two enriched KGs for three relation prediction tasks: PPI prediction, GDA prediction, and disease prediction. 
Each benchmark dataset comprises several pairs of biomedical entities (or instances) that can be of the same type (protein-protein) or distinct types (gene-disease and disease-patient) with the respective label 
(1 for the positive pairs and zero for the negative pairs).
Tables ~\ref{tab:statistics2} and \ref{tab:statistics} show the KGs' and datasets' statistics for each task.
Since for GDA prediction and disease prediction, the target relation happens between two types of instances (genes and diseases for GDA prediction and diseases and patients for disease prediction), the instance numbers in Table~\ref{tab:statistics} appear separately.
Moreover, in the case of PPI prediction, we exclusively employ the GO KG that has been subjected to a negative statement enrichment approach. However, when it comes to GDA prediction and disease prediction, we rely on the HP KG, which lacks a negative statement enrichment approach, resulting in a significant imbalance between the number of positive and negative statements.

\begin{table}[!tb]
\centering
\setlength{\tabcolsep}{4pt}
\small
\caption{Statistics for each ontology regarding classes, nodes, edges.}
\begin{tabular}{lrrrrr}
\toprule
& GO & HP \\ \midrule
Classes & 50918 & 17060 \\
Literals and blank nodes & 532373 & 442246\\
Edges & 1425102 &  1082859 \\
\bottomrule
\end{tabular}
\label{tab:statistics2} 
\end{table}
  
\begin{table}[!tb]
\centering
\setlength{\tabcolsep}{4pt}
\small
\caption{Statistics for each task's dataset regarding the number of instances, pairs, positive and negative statements.}
\begin{tabular}{lrrrrr}
\toprule
& PPI prediction  &  & GDA prediction & & Disease Prediction \\ \cmidrule{2-2} \cmidrule{4-4} \cmidrule{6-6}%\midrule
Instances & 440 && 174 + 107 && 1033 + 660 \\
Positive Pairs & 1024 && 107 && 660   \\
Negative Pairs & 1024 && 107 && 681120  \\
Positive statements& 7364 &&  14828 & & 38130\\
Negative statements & 8579 && 9191 & & 179 \\
\bottomrule
\end{tabular}
\label{tab:statistics} 
\end{table}

\begin{figure}[!tb]
    \centering
    \includegraphics[width=\textwidth]{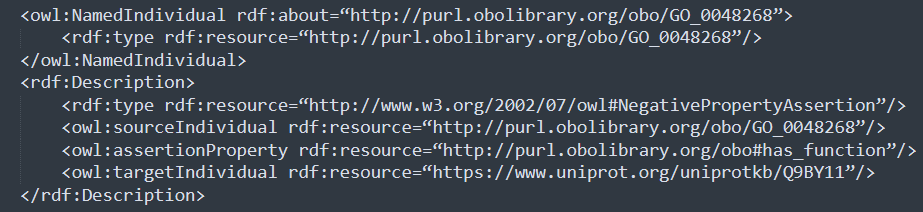}
    \caption{Example of how the negative statements are defined in the OWL file.}
    \label{fig:negativestatement}
\end{figure}

To build these datasets, we adopt three main steps. The first one consists of enriching the KGs. 
The KG is constructed using the owlready2 package\footnote{https://owlready2.readthedocs.io/en/v0.37/}, which parses the ontology file in OWL format and processes the annotation file. The annotation file contains positive and negative statements used to describe entities.
We use the guidelines established by the W3C\footnote{https://www.w3.org/TR/owl2-mapping-to-rdf/} to define the negative statements as negative object property assertions\footnote{https://www.w3.org/TR/owl2-syntax/\#Negative\_Object\_Property\_Assertions}.
To do so, we use metamodeling and represent each ontology class as a class and an individual. This situation translates into using the same IRI. Then, we use a negative object property assertion to state that the individual representing a biomedical entity is not connected by the object property expression to the individual representing an ontology class, as depicted in Figure~\ref{fig:negativestatement}. The second step consists of extracting pairs of entities from bioinformatic databases. The third step involves selecting the pairs containing KG entities that are well described with positive and negative statements. 
% \begin{lstlisting}[language=XML]
% <owl:NamedIndividual rdf:about=“http://purl.obolibrary.org/obo/GO_0048268”>
%     <rdf:type rdf:resource=“http://purl.obolibrary.org/obo/GO_0048268”/>
% </owl:NamedIndividual>
% <rdf:Description>
%     <rdf:type rdf:resource=“http://www.w3.org/2002/07/owl#NegativePropertyAssertion”/>
%       <owl:sourceIndividual rdf:resource=“http://purl.obolibrary.org/obo/GO_0048268”/>
%       <owl:assertionProperty rdf:resource=“http://purl.obolibrary.org/obo/go.owl#has_function”/>
%       <owl:targetIndividual rdf:resource=“https://www.uniprot.org/uniprotkb/Q9BY11”/>
% </rdf:Description>
% \end{lstlisting}

The following subsections describe in more detail the KGs as the characteristics of each task. 

%%%%%%%%%%%%%%%%%%%%%%%%%%%%%%%%%%% Biomedical KGs
\subsection{Biomedical Knowledge Graphs}

Two KGs back-boned by biomedical ontologies are used: the GO KG and the HP KG. Table~\ref{tab:statistics2} shows the statistics for each ontology. 

The GO is used to describe gene products (proteins or genes) according to the molecular functions they perform, the biological processes they are involved in, and the cellular components where they act. 
The GO KG is built by integrating three sources: the GO\footnote{The GO was downloaded on September 2021. It is available at~\url{http://release.geneontology.org/2021-09-01/ontology/index.html}} itself, the GO Annotation data\footnote{The GO positive annotations were downloaded on January 2021. It is available at~\url{http://release.geneontology.org/2021-01-01/annotations/index.html}.}~\cite{gene2021gene}, and negative GO associations produced in~\cite{vesztrocy2020benchmarking}\footnote{The negative annotations were downloaded from \url{https://lab.dessimoz.org/20\_not}}.

A GO annotation links a specific gene product with a particular GO class. 
%They link Uniprot protein identifiers with the GO classes that describe them. 
The majority of GO annotation data corresponds to positive statements. However, the GO annotation has the qualifier ‘NOT’ for a few cases, meaning that a gene product has been proven not to carry out a specific function. The annotations that possess this qualifier were added as negative statements. 
In addition to these negative statements, the GO KG was also enriched with negative statements derived from expert-curated annotations of protein families on phylogenetic trees. 
The idea is that, if no evidence exists to suggest otherwise, gene function is maintained over time through evolution.
Therefore, after expert curators have annotated ancestral states in gene phylogenies with GO classes, they check if the annotations are propagated down the phylogeny. When there is evidence that the function is absent in a specific sub-tree, a negative statement is added to that protein.
These enriched negative statements were filtered so there were no contradictions with the GO annotation data. 

HP characterizes phenotypic abnormalities discovered in human hereditary diseases according to five semantic aspects: phenotypic abnormalities, mode of inheritance, clinical course, clinical modifier and frequency. HP annotations can link diseases, patients or genes to HP classes via positive and negative statements. The construction of HP KG\footnote{The HP was downloaded on October 2022, while the HP annotations were downloaded on November 2021. A link to these versions is no longer available.} is similar to that of the GO KG. A negative annotation from HP that includes 'NOT' indicates that a disease does not cause that phenotype, so they are included as negative statements.

%%%%%%%%%%%%%%%%%%%%%%%%%%%%%%%%%%% PPI Dataset
\subsection{Protein-Protein Interaction Prediction Dataset}

Predicting PPIs is a fundamental task in molecular biology for understanding biological systems. Given the high cost of experimentally determining PPI, many computational approaches for PPI prediction based on available functional information described by the GO~\cite{GO2018} have been proposed to find protein pairs likely to interact and thus provide a selection of good candidates for experimental analysis. Therefore, the GO KG is used to describe the proteins of the dataset.

The positive examples are extracted from the STRING~\cite{STRING2021} database. Our selection of protein pairs was based on the following criteria: (i) interactions between proteins had to be curated or experimentally determined rather than computationally determined;  (ii) interactions needed to have a confidence score above 0.950 to ensure high confidence; 
% (iii) each protein must have at least one positive GO association and one negative GO association. 
(iii) each protein must have at least one positive statement for a GO class and one negative statement for another GO class.
The negative examples are generated by random negative sampling over the set of proteins of the positive examples.
% A final annotation file with positive and negative annotations was generated for the proteins in the dataset, with a "+" or "-" indicating whether they are positive or negative.

%%%%%%%%%%%%%%%%%%%%%%%%%%%%%%%%%%% Gene-Disease Dataset
\subsection{Gene-Disease Association Prediction Dataset}

Knowing which genes are associated with a specific disease is crucial to understanding the disease mechanisms and recognising potential biomarkers or therapeutic targets. 
However, once again, validating these associations in the wet lab is expensive and time-consuming.
This has prompted the evolution of computational methods to identify the most promising associations to be further validated.

The two KGs are used for the GDA prediction task dataset. GO KG describes the genes, and HP KG describes the diseases. 
The target relations to predict are extracted from \mbox{DisGeNET}~\cite{pinero2019disgenet}. Adapting the approach described in~\cite{nunes2021predicting}, we considered the following criteria to select gene-disease pairs: (i) each gene must have at least one positive statement for a GO class and one negative statement for another GO class; (ii) each disease must have at least one positive statement for an HP class and one negative statement for an HP class. 
We sampled random negative examples of the same genes and diseases to create a balanced dataset. 
%Once again, we downloaded the positive and negative annotations files and then combined them into a final disease annotations file.

%%%%%%%%%%%%%%%%%%%%%%%%%%%%%%%%%%% Disease-Patient Dataset
\subsection{Disease Prediction Datasets}

Since human diseases are a complex phenomenon, disease prediction is an essential but still complicated task that must be executed accurately and efficiently. 
Therefore, using computational methods to help physicians prioritize diseases is highly advantageous.

The dataset to predict if a synthetic patient has been diagnosed with a specific disease is generated by adapting the methodology proposed in~\cite{masino2014clinical}.
Thirty-three mendelian diseases for which they knew the penetrance of each phenotype are selected. Penetrance indicates the likelihood that a patient suffering from a specific disease will exhibit a particular phenotype.
For each of these 33 diseases, 20 synthetic patients diagnosed with that disease are created.  
The patients' positive annotation is determined by the disease's penetrance and the patient's gender. The gender is defined randomly with an equal likelihood for both genders.
For example, the 'Aarskog-Scott syndrome' is annotated with the phenotype 'Ptosis' with a penetrance of 0.5061, meaning that approximately half of the synthetic patients diagnosed with that disease will have a positive statement for this phenotype.
The negation of phenotypes does not have a penetrance associated, so synthetic patients inherit the negative phenotypes related to the disease. For example, since the disease 'Aarskog-Scott syndrome' is annotated with 'NOT Decreased Fertility', each patient will have a negative statement for this phenotype.
Furthermore, 1000 diseases were randomly chosen to add complexity to the task. These diseases are annotated with positive and negative statements.

Random annotations can also be added to patients to emulate a more realistic situation where a patient is associated with phenotypes unrelated to the patient's disease. In addition to the disease prediction dataset, we present three versions with random annotations. The number of random annotations is defined by a percentage Noi (Noi=[0, 0.1,0.2,0.4]) concerning a given patient's total number of annotations. For example, if Noi=0.5, half of the full annotations of a given patient are added.
Table~\ref{tab:statistics3} shows the number of positive and negative statements for each noise version.

\begin{table}[!tb]
\centering
\setlength{\tabcolsep}{4pt}
\small
\caption{Statistics for each noise version regarding the number of positive and negative statements.}
\begin{tabular}{rrr}
\toprule
\textbf{Noise} &  \textbf{Positive Statements} & \textbf{Negative Statements} \\  \midrule
0.1 & 40592 & 216 \\
0.2 & 37195 &  214 \\
0.4 & 38242 &  192\\
\bottomrule
\end{tabular}
\label{tab:statistics3} 
\end{table}

%%%%%%%%%%%%%%%%%%%%%%%%%%%%%%%%%%%%%%%%%%%%%%%%%%%%%%% Validation
\section{Validation of the Datasets}

KG embedding methods~\cite{wang2017knowledge} have been successfully employed in several biomedical applications~\cite{wang2017knowledge}. 
Since these methods map KGs into low-dimensional spaces, they have emerged as a popular way to generate features for machine learning tasks. Therefore, we use two KG embedding methods to evaluate our datasets - RDF2Vec~\cite{ristoski2016rdf2vec} and OWL2Vec*~\cite{chen2021owl2vec}. 
RDF2Vec is a path-based method that generates random walks in the KG that constitutes the corpus of word sequences given as input to a neural language model. OWL2Vec* was designed to learn ontology embeddings and it also employs direct walks on the graph to learn graph structure.
These embedding methods generate representations of the biomedical entities that are combined using the binary Hadamard operator to represent the pair. 

The pair representations are then fed into a Random Forest algorithm for training using Monte Carlo cross-validation (MCCV)~\cite{XU20011}. MCCV is a variation of traditional $k$-fold cross-validation in which the data is divided into training and testing sets (with $\beta$ being the proportion of the dataset to include in the test split) $M$ times. 
Our experiments use MCCV with $M=30$ and $\beta=0.3$ for PPI and GDA prediction. Given the large number of pairs for disease prediction, we use MCCV with $M=5$ and $\beta=0.3$. 

Each embedding method is run with two different KGs, one with only positive statements and the other with both positive and negative statements.
Table~\ref{tab:Prediction} reports each task's median of recall, precision and weighted average F-measure. 

\begin{table*}[!tb]
\small
\centering
\caption{Median precision, recall and weighted average F-measure (Pr, Re, and F1) for PPI, GDA, and disease prediction using only positive statements (Pos) or positive and negative statements (Pos+Neg).}
\begin{tabular}{llrrrrrrrrrrr}
\toprule
\multirow{2}{*}{\textbf{Method}} & \multirow{2}{*}{\textbf{Statements}}  & \multicolumn{3}{c}{\textbf{PPI}} && \multicolumn{3}{c}{\textbf{GDA}}&& \multicolumn{3}{c}{\textbf{Disease}}\\
\cmidrule{3-5} \cmidrule{7-9} \cmidrule{11-13}
& & \textbf{Pr}   & \textbf{Re} & \textbf{F1} & & \textbf{Pr}   & \textbf{Re} & \textbf{F1} & & \textbf{Pr}   & \textbf{Re} & \textbf{F1}\\ \midrule
\multirow{2}{*}{RDF2Vec} 
& Pos    & 0.831   & 0.826                & 0.828    &  & 0.623                & 0.625          & 0.615     &  &   0.994 & 0.742 & 0.850  \\
& Pos+Neg     & 0.847  & 0.844         & 0.845  &  &  0.654                & 0.600          & 0.645        & &  1.000 & 0.771 & 0.870  \\

 \midrule
\multirow{2}{*}{OWL2Vec*} 
& Pos   & 0.833   & 0.806                & 0.823    &  & 0.652                & 0.656 & 0.646     &  &  0.975 & 0.584 &  0.730\\
& Pos+Neg    & 0.860  & 0.812           & 0.840  &  &    0.625                & 0.661 & 0.630    &  &  0.980 & 0.563 &  0.713 \\

\bottomrule
\end{tabular}
\label{tab:Prediction}
\end{table*}

\begin{figure}[!t]
\begin{minipage}{.5\linewidth}
\centering
\subfloat[PPI prediction]{\label{main:a}\includegraphics[scale=.41]{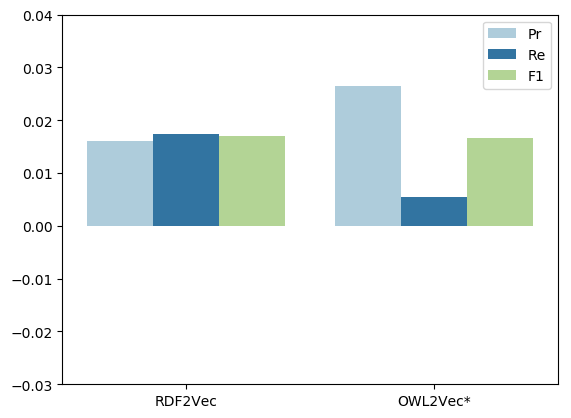}}
\end{minipage}%
\begin{minipage}{.5\linewidth}
\centering
\subfloat[GDA prediction]{\label{main:b}\includegraphics[scale=.41]{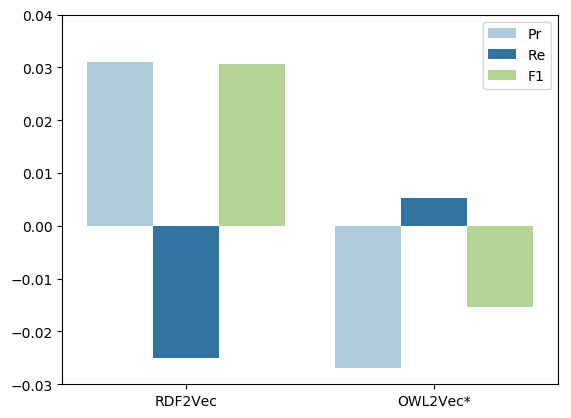}}
\end{minipage}\par\medskip
\centering
\subfloat[Disease prediction]{\label{main:c}\includegraphics[scale=.41]{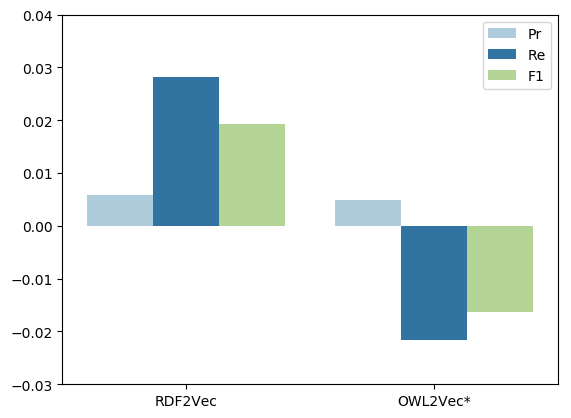}}
\caption{Barplots showing the differences in precision, recall, weighted average F-measure (Pr, Re, and F1) between using positive and negative statements or only positive statements.}
\label{fig:differencespos-neg}
\end{figure}

Figure~\ref{fig:differencespos-neg} compares the impact of using only positive statements versus both positive and negative statements on our datasets. The bars represent the difference in performance for precision, recall and weighted average F-measure, with upward bars indicating improved performance with both positive and negative statements and downward bars indicating decreased performance.

% The experiments show that the added information given by negative statements generally improves the performance of most KG embedding methods.
The experiments show that the added information given by negative statements generally improves the performance of RDF2Vec. However, for OWL2Vec*, the performance only improves for PPI prediction.

%%%%%%%%%%%%%%%%%%%%%%%%%%%%%%%%%%%%%%%%%%%%%%%%%%%%%%% USAGE
\section{Using the Benchmark}

%All datasets will be available on Zenodo under a CC BY 4.0 license.
All datasets are available on Zenodo\footnote{https://doi.org/10.5281/zenodo.7709195} under a CC BY 4.0 license.
For each dataset, we provide access to two types of files: (1) one TSV file containing pairs of entities and information about whether a relationship exists between them or not; (2) OWL files containing the KG used to describe the biomedical entities that appear in the TSV file. 
Together, these files can be used to perform relation prediction tasks since the TSV file provides the specific entities and relations that need to be predicted, while the OWL file provides the necessary background knowledge for generating the features.

%%%%%%%%%%%%%%%%%%%%%%%%%%%%%%%%%%%%%%%%%%%%%%%%%%%%%%% CONCLUSIONS
\section{Conclusions}

Benchmark datasets are essential for evaluating and comparing the performance of different approaches that work over KGs. This paper presents a collection of datasets for three relation prediction tasks in the biomedical domain: PPI prediction, GDA prediction, and disease prediction. 
The biomedical domain is chosen since it is already demonstrated that the inadequacy of approaches to take into consideration negative statements is a limitation for several biomedical applications.
However, although the datasets are domain-specific, they can be used to evaluate approaches outside the biomedical domain. 

The datasets are validated using two popular KG embedding methods to generate features that are then given as input for a classifier. The results highlight the importance of incorporating negative statements into KGs to create more accurate representations of KG entities.

%%%%%%%%%%%%%%%%%%%%%%% ACKNOWLEDGMENTS
\subsubsection*{Acknowledgements}

C. P., S. S., R. T. S. are funded by the FCT through LASIGE Research Unit (ref. UIDB/00408/2020 and ref. UIDP/00408/2020), and the FCT PhD grant (ref. SFRH/BD/145377/2019). It was also partially supported by the KATY project, which has received funding from the European Union’s Horizon 2020 research and innovation programme under grant agreement No 101017453, and by HfPT: Health from Portugal under the Portuguese Plano de Recuperação e Resiliência.
The authors thank Lina Aveiro for the preliminary results of this work.

\bibliographystyle{splncs04}
\bibliography{references}
\end{document}